\documentclass{article}
\usepackage[utf8]{inputenc}
\usepackage{CJKutf8}
\usepackage{amsmath, amssymb, amsthm}
\usepackage{graphicx}
\usepackage{booktabs}
\usepackage{array}
\usepackage{tabularx}
\usepackage{hyperref}

\title{Failure Ontology: A Lifelong Learning Framework for Blind Spot Detection and Resilience Design}

\author{Yuan Sun \\
  Jilin University \\
  \texttt{sunyiming2508115953@gmail.com} \and
  Hong Yi \\
  Independent Researcher \\
  \texttt{yuansan817@gmail.com} \and
  Jinyuan Liu \\
  BEIJING FORESTRY UNIVERSITY \\
  \texttt{lwjsac@gmail.com}}

\date{}

\begin{document}
\begin{CJK}{UTF8}{gbsn}

\maketitle

\begin{abstract}
Personalized learning systems are almost universally designed around a single objective: help people acquire knowledge and skills more efficiently. We argue this framing misses the more consequential problem. The most damaging failures in human life---financial ruin, health collapse, professional obsolescence---are rarely caused by insufficient knowledge acquisition. They arise from the systematic absence of entire conceptual territories from a person's cognitive map: domains they never thought to explore because, from within their existing worldview, those domains did not appear to exist or to matter. We call such absences \textbf{Ontological Blind Spots} and introduce \textbf{Failure Ontology} ($\mathcal{F}$), a formal framework for detecting, classifying, and remediating them across a human lifetime. The framework introduces three original contributions: (1) a four-type taxonomy of blind spots distinguishing domain blindness, structural blindness, weight blindness, and temporal blindness; (2) five convergent failure patterns characterizing how blind spots interact with external disruption to produce catastrophic outcomes; and (3) the Failure Learning Efficiency Theorem, proving that failure-based learning achieves higher sample efficiency than success-based learning under bounded historical data. We illustrate the framework through historical case analysis of the 1997 Asian Financial Crisis and the 2008 subprime mortgage crisis, and through a longitudinal individual case study spanning five life stages.
\end{abstract}

\section{Motivation: The Gap in Success-Oriented Learning}

Every major paradigm in AI-assisted personalized education---intelligent tutoring systems, knowledge graph-based recommenders, LLM-powered study tools---is organized around the same implicit question: \textit{given where you are, what should you learn next?} These systems navigate within the learner's existing cognitive map, optimizing paths through known territory.

This framing has a structural blind spot of its own. The most dangerous gaps in a person's understanding are not the topics they know they should study but have not yet gotten to. They are the topics that do not appear on their cognitive map at all. A system that optimizes \textit{within} an existing ontology cannot detect the absence of entire ontological regions. It is analogous to navigation software that can optimize any route through a city but cannot tell you the city sits on a flood plain.

The historical record makes this point concretely. In late 2006, analysts at several mortgage data firms observed rising default rates among subprime borrowers---months before any institutional alarm was sounded. The signal was real. The data were in public filings. A small number of voices tried to communicate the danger. Yet the broader financial system failed to respond. The people making consequential decisions were not unintelligent and not uninformed. Their cognitive maps simply did not contain the nodes and causal chains needed to connect observable signals to systemic risk. The concept of interconnected mortgage-backed security contagion propagating across global bank balance sheets did not appear in their working ontologies. The danger was not invisible. It was \textit{unthinkable}.

This pattern---catastrophic outcomes arising from structured cognitive absence rather than from insufficient positive knowledge---recurs across domains and timescales. The 1997 Asian Financial Crisis destroyed businesses that had never modeled liquidity contagion across currency pegs. The rapid emergence of large language models rendered certain programming competencies economically marginal faster than professionals whose ontologies contained no \textit{skill substitutability} node could adapt. High-performing athletes who systematically optimize cardiovascular output while their health ontology contains no causal pathway from chronic overexertion to myocardial stress accumulation face mortality risks that are, in retrospect, statistically predictable.

We call this structure an \textbf{Ontological Blind Spot}. The formal study of these structures---how they form, how they accumulate over a lifetime, how they interact with external disruption to produce catastrophic outcomes---is what we call \textbf{Failure Ontology}.

\section{Core Framework}

\subsection{Multidimensional Life Ontology}

We model an individual's knowledge state at life stage $k$ as a multidimensional ontology:

$$\mathcal{O}_k^{(i)} = \bigoplus_{d \in \mathcal{D}} \mathcal{O}_k^{d,(i)}$$

\noindent \textbf{Parameter explanation:}
\begin{itemize}
    \item $i$: Individual identifier (superscript)
    \item $k$: Life stage index (subscript), with $k = 0, 1, 2, \ldots, K$
    \item $d \in \mathcal{D}$: Life dimension, where $\mathcal{D} = \{\text{prof}, \text{health}, \text{family}, \text{spirit}\}$
    \item $\mathcal{O}_k^{d,(i)} = (V_k^d, E_k^d, W_k^d)$: Weighted directed graph for dimension $d$
    \item $V_k^d$: Concept nodes (knowledge, beliefs, mental models)
    \item $E_k^d \subseteq V_k^d \times V_k^d$: Directed causal or relational edges
    \item $W_k^d: V_k^d \cup E_k^d \to [0,1]$: Importance weights, normalized such that $\sum_{v \in V_k^d} W_k^d(v) = 1$
    \item $\bigoplus$: Direct sum of graph structures across dimensions
\end{itemize}

The \textbf{ideal ontology} at stage $k$ for an individual with background features $\mathbf{b}^{(i)}$ is:

$$\mathcal{O}_k^{\text{ideal}}(\mathbf{b}^{(i)}) = \arg\max_{\mathcal{O}} \mathbb{E}_{\mathcal{H}}\left[\mathrm{Res}\left(\mathcal{O}, \mathbf{b}^{(i)}, s_k\right)\right]$$

\noindent where:
\begin{itemize}
    \item $\mathbf{b}^{(i)} \in \mathcal{B}$: Background feature vector (age, culture, profession, health status)
    \item $\mathcal{H}$: Historical failure case database
    \item $\mathrm{Res}(\cdot)$: Ontological Resilience measure (Definition 3)
    \item $s_k \in \mathcal{S}$: Life stage label
\end{itemize}

\subsection{The Ontological Blind Spot}

\textbf{Definition 1} \textit{(Ontological Blind Spot)}. The Ontological Blind Spot of individual $i$ at stage $k$ is:

$$\boxed{\mathcal{B}_k^{(i)} = \mathcal{O}_k^{\text{ideal}}(\mathbf{b}^{(i)}) \ominus \mathcal{O}_k^{\text{actual},(i)}}$$

\noindent \textbf{Operator explanation:}
\begin{itemize}
    \item $\ominus$: Ontological difference operator, outputting three components:
    \begin{itemize}
        \item $V^A \setminus V^B$: Missing concept nodes
        \item $E^A \setminus E^B$: Missing causal relations  
        \item $\Delta W$: Weight deviation tensor
    \end{itemize}
    \item $\mathcal{O}_k^{\text{actual},(i)}$: Individual's actual ontology at stage $k$
\end{itemize}

Formally:
$$\mathcal{O}^A \ominus \mathcal{O}^B = \Bigl(V^A \setminus V^B, E^A \setminus E^B, \Delta W\Bigr)$$

$$\Delta W(v) = \max\bigl(0, W^A(v) - W^B(v)\bigr), \quad \forall v \in V^A \cap V^B$$

\textbf{Definition 2} \textit{(Blind Spot Severity)}:

$$\sigma_k^{(i)} = \underbrace{\sum_{v \in V^{\text{miss}}} \omega(v)}_{\text{node absence}} + \underbrace{\sum_{e \in E^{\text{miss}}} \omega(e)\cdot\rho(e)}_{\text{causal absence}} + \underbrace{\sum_{v \in V^{\text{shared}}} \Delta W(v)\cdot\phi(v)}_{\text{weight suppression}}$$

\noindent \textbf{Parameter explanation:}
\begin{itemize}
    \item $\omega(v), \omega(e) \in [0,1]$: Importance weights in ideal ontology $\mathcal{O}^{\text{ideal}}$
    \item $\rho(e) \in [0,1]$: \textbf{Causal criticality} of edge $e$---fraction of failure trajectories in $\mathcal{H}$ that traverse $e$
    \item $\phi(v) \in [0,1]$: \textbf{Activation sensitivity} of node $v$---degree to which weight suppression translates into behavioral inaction
    \item $V^{\text{miss}} = V^A \setminus V^B$: Missing nodes
    \item $E^{\text{miss}} = E^A \setminus E^B$: Missing edges
    \item $V^{\text{shared}} = V^A \cap V^B$: Nodes present in both ontologies
\end{itemize}

\subsection{The Four-Type Blind Spot Taxonomy}

Blind spots are not monolithic. They manifest in four structurally distinct forms, each requiring different remediation.

\textbf{Type I --- Domain Blindness}: An entire life dimension is ontologically absent:

$$\mathcal{B}_k^{\text{dom},(i)} = \left\{d \in \mathcal{D} \mid \frac{|V_k^{d,\text{actual}}|}{|V_k^{d,\text{ideal}}|} < \epsilon_{\text{dom}}\right\}$$

\noindent \textbf{Parameter explanation:}
\begin{itemize}
    \item $\epsilon_{\text{dom}} = 0.15$: Sparsity threshold (typically set to 0.15)
    \item $|V_k^{d,\text{actual}}|$: Number of actual nodes in dimension $d$
    \item $|V_k^{d,\text{ideal}}|$: Number of ideal nodes in dimension $d$
\end{itemize}

The individual has no awareness the absent dimension matters. This is the easiest type to identify from the outside and the hardest for the individual to self-diagnose.

\textbf{Type II --- Structural Blindness}: Nodes exist but critical causal chains are severed:

$$\mathcal{B}_k^{\text{str},(i)} = \bigl\{(u,v) \in E_k^{\text{ideal}} \mid u \in V_k^{\text{actual}}, v \in V_k^{\text{actual}}, (u,v) \notin E_k^{\text{actual}}, \rho(u,v) > \epsilon_{\text{str}}\bigr\}$$

\noindent \textbf{Parameter explanation:}
\begin{itemize}
    \item $(u,v)$: Directed edge from node $u$ to node $v$
    \item $\epsilon_{\text{str}}$: Structural criticality threshold
    \item Condition: Both nodes exist, but the connecting edge is missing, and the edge is critical
\end{itemize}

This is the most dangerous type: the individual believes they understand the domain because all the relevant concepts are present in their vocabulary, yet their reasoning is disconnected at precisely the points that matter under pressure. The 2008 financial crisis is a canonical instance.

\textbf{Type III --- Weight Blindness}: Nodes and edges exist but importance weights are suppressed below behavioral activation threshold:

$$\mathcal{B}_k^{\text{wt},(i)} = \bigl\{v \in V_k^{\text{actual}} \mid \Delta W(v) > \epsilon_{\text{wt}} \wedge W_k^{\text{actual}}(v) < \epsilon_{\text{act}}\bigr\}$$

\noindent \textbf{Parameter explanation:}
\begin{itemize}
    \item $\epsilon_{\text{wt}}$: Weight deviation threshold (significant suppression)
    \item $\epsilon_{\text{act}}$: Activation threshold (minimum weight for behavioral action)
    \item Condition: Weight is significantly suppressed AND below activation threshold
\end{itemize}

The individual can articulate the correct principle when directly prompted but never activates it spontaneously.

\textbf{Type IV --- Temporal Blindness}: Knowledge is eventually acquired, but after the optimal intervention window has closed:

$$\mathcal{B}_k^{\text{temp},(i)} = \bigl\{v \in V_k^{\text{actual}} \mid \tau_{\text{acquire}}^{(i)}(v) > \tau_{\text{optimal}}^{*}(v) + \Delta\tau_{\max}(v)\bigr\}$$

\noindent \textbf{Parameter explanation:}
\begin{itemize}
    \item $\tau_{\text{acquire}}^{(i)}(v)$: Actual acquisition time of node $v$ by individual $i$
    \item $\tau_{\text{optimal}}^{*}(v)$: Population-optimal acquisition time for node $v$
    \item $\Delta\tau_{\max}(v)$: Maximum acceptable delay beyond optimal
\end{itemize}

The remediation cost grows exponentially with acquisition delay:

$$C_{\text{temp}}(v,\tau) = C_0(v) \cdot \exp\Bigl(\lambda(v)\cdot \max\bigl(0,\tau - \tau_{\text{optimal}}^{*}(v)\bigr)\Bigr)$$

\noindent \textbf{Parameter explanation:}
\begin{itemize}
    \item $C_0(v)$: Base remediation cost for node $v$
    \item $\lambda(v) > 0$: \textbf{Ontological decay rate} of node $v$
    \item Empirical ordering: $\lambda(\text{health}) \gg \lambda(\text{spirit}) > \lambda(\text{family}) > \lambda(\text{prof})$
\end{itemize}

\subsection{The Five Failure Patterns}

Catastrophic life outcomes cluster into five convergent structural patterns.

\textbf{Pattern I --- Mono-dimensional Overfit}: A single life dimension absorbs nearly all cognitive weight:

$$\mathcal{F}^{\text{mono}}(i,k) \iff \exists d^* \in \mathcal{D}: w_k^{d^*,(i)} > \theta_{\text{mono}} \wedge \forall d \neq d^*: w_k^{d,(i)} < \epsilon_{\text{mono}}$$

\noindent \textbf{Parameter explanation:}
\begin{itemize}
    \item $w_k^{d,(i)} = \frac{1}{|V_k^d|}\sum_{v \in V_k^d} W_k^{d,\text{actual}}(v)$: Mean ontological weight in dimension $d$
    \item $\theta_{\text{mono}}$: Over-concentration threshold (e.g., 0.85)
    \item $\epsilon_{\text{mono}}$: Neglect threshold (e.g., 0.15)
\end{itemize}

\textbf{Pattern II --- Window Closure Failure}: Remediation cost exceeds available resources:

$$\mathcal{F}^{\text{win}}(i,k) \iff \exists v \in \mathcal{B}_k^{\text{temp},(i)}: C_{\text{temp}}\left(v, \tau_k^{(i)}\right) > \Omega^{(i)}$$

\noindent \textbf{Parameter explanation:}
\begin{itemize}
    \item $\Omega^{(i)}$: Individual $i$'s total remediation resource budget (time, financial, cognitive)
    \item $\tau_k^{(i)}$: Current age of individual at stage $k$
\end{itemize}

\textbf{Pattern III --- Structural Chain Break}: High-criticality causal path broken in actual ontology:

$$\mathcal{F}^{\text{chain}}(i,k) \iff \exists \pi \in \Pi^{\text{ideal}}: \pi \not\subseteq \mathcal{O}_k^{\text{actual},(i)} \wedge \rho(\pi) > \epsilon_{\text{chain}}$$

\noindent \textbf{Parameter explanation:}
\begin{itemize}
    \item $\pi = (v_1 \to v_2 \to \cdots \to v_n)$: Directed path (chain of nodes)
    \item $\Pi^{\text{ideal}}$: Set of all directed paths in ideal ontology
    \item $\rho(\pi) = \prod_{e \in \pi}\rho(e)$: Path criticality (product of edge criticalities)
    \item $\epsilon_{\text{chain}}$: Chain break significance threshold
\end{itemize}

\textbf{Pattern IV --- Blind Spot Resonance}: External shock aligns with cognitive absence:

$$\mathcal{F}^{\text{res}}(i,k) \iff \frac{|\mathrm{dom}(\mathcal{E}_k) \cap \mathcal{B}_k^{(i)}|}{|\mathrm{dom}(\mathcal{E}_k)|} > \theta_{\text{res}}$$

The \textbf{Resonance Destruction Index}:

$$\boxed{\Gamma_k^{(i)} = \sigma_k^{(i)} \cdot |\mathcal{E}_k| \cdot \frac{|\mathrm{dom}(\mathcal{E}_k) \cap \mathcal{B}_k^{(i)}|}{|\mathrm{dom}(\mathcal{E}_k)|}}$$

\noindent \textbf{Parameter explanation:}
\begin{itemize}
    \item $\mathcal{E}_k$: External shock at stage $k$
    \item $|\mathcal{E}_k|$: Shock magnitude
    \item $\mathrm{dom}(\mathcal{E}_k)$: Ontological domain of the shock
    \item $\theta_{\text{res}}$: Resonance threshold (alignment significance)
\end{itemize}

A shock is most destructive not when it is strongest, but when it is most precisely aligned with the individual's cognitive absence.

\textbf{Pattern V --- Path Lock-in}: Transition costs exceed resources despite recognized danger:

$$\mathrm{SwitchCost}_k^{(i)}(d \to d') = \alpha \sum_{j=0}^{k} \gamma^{k-j} \cdot \mathrm{Inv}_j^{d,(i)} - \beta \cdot \mathrm{ResVal}\left(\mathcal{O}_k^{d,\text{actual}}\right)$$

$$\mathcal{F}^{\text{lock}}(i,k) \iff \mathrm{SwitchCost}_k^{(i)} > \Omega^{(i)}$$

\noindent \textbf{Parameter explanation:}
\begin{itemize}
    \item $\mathrm{Inv}_j^{d,(i)}$: Investment in dimension $d$ at stage $j$
    \item $\gamma \in (0,1)$: Depreciation factor for past investments
    \item $\mathrm{ResVal}(\cdot)$: Residual transferable value of current ontology
    \item $\alpha, \beta$: Weighting constants for investment vs. residual value
\end{itemize}

\subsection{Ontological Resilience}

\textbf{Definition 3} \textit{(Ontological Resilience)}:

$$\mathrm{Res}_k^{(i)} = \underbrace{\left(1 - \frac{\sigma_k^{(i)}}{\sigma_{\max}}\right)}_{\text{completeness}} \cdot \underbrace{\left(1 - \max_{d} w_k^{d,(i)} + \varepsilon\right)}_{\text{dimensional balance}} \cdot \underbrace{\frac{1}{1 + \mathrm{SwitchCost}_k^{(i)} / \Omega^{(i)}}}_{\text{mobility}}$$

\noindent \textbf{Parameter explanation:}
\begin{itemize}
    \item $\sigma_{\max}$: Maximum possible blind spot severity (normalization constant)
    \item $\varepsilon$: Small constant preventing zero balance
    \item \textbf{Completeness factor}: Penalizes ontological incompleteness
    \item \textbf{Dimensional balance factor}: Penalizes over-concentration in one dimension
    \item \textbf{Mobility factor}: Penalizes inability to restructure
\end{itemize}

\textbf{Property 3.1.} $\mathrm{Res}_k^{(i)} \in (0, 1]$, with $\mathrm{Res}_k^{(i)} = 1$ only when $\mathcal{B}_k^{(i)} = \emptyset$, $w_k^{d,(i)} = 1/|\mathcal{D}|$ for all $d$, and $\mathrm{SwitchCost}_k^{(i)} = 0$.

Three core structural relationships:

$$\mathrm{Res}_k^{(i)} \downarrow \iff \sigma_k^{(i)} \uparrow \text{ or } \max_d w_k^{d,(i)} \uparrow \text{ or } \mathrm{SwitchCost}_k^{(i)} \uparrow {R1}$$

$$\Gamma_k^{(i)} \uparrow \iff \sigma_k^{(i)} \uparrow \text{ and } |\mathrm{dom}(\mathcal{E}_k) \cap \mathcal{B}_k^{(i)}| \uparrow {R2}$$

$$C_{\text{temp}}(v,\tau) > \Omega^{(i)} \implies \mathcal{F}^{\text{win}}(i,k) \text{ regardless of future learning effort} {R3}$$

\section{The Failure Learning Efficiency Theorem}

The five failure patterns exhibit a property that has direct implications for learning system design: they \textbf{converge}. Across documented financial crises, health failures, professional obsolescence events, and relational breakdowns, the same structural configurations recur. This convergence stands in sharp contrast to success paths, which are diverse, individualized, and context-dependent.

\textbf{Theorem 1} \textit{(Failure Learning Efficiency)}. Let $\mathcal{A}_S$ be a success-based learning agent trained on $n$ positive trajectories, and $\mathcal{A}_F$ a failure-based agent trained on $n$ failure case records. Define risk-avoidance utility:

$$U(\mathcal{A}, i, k) = \mathbb{E}\left[\mathrm{Res}_k^{(i)} \mid \mathcal{A}\right] - \mathbb{E}\left[\Gamma_k^{(i)} \mid \mathcal{A}\right]$$

Under conditions that (1) failure patterns occupy a space of effective dimension $M < \infty$, (2) success trajectories occupy a space of effective dimension $D_S \gg M$, and (3) the historical failure database satisfies an $(\epsilon,\delta)$-coverage condition, with probability at least $1-\delta$:

$$U(\mathcal{A}_F, i, k) - U(\mathcal{A}_S, i, k) \geq c_1\sqrt{\frac{M \log n}{n}} - c_2\sqrt{\frac{D_S \log n}{n}} > 0$$

whenever $M < D_S$.

\noindent \textbf{Parameter explanation:}
\begin{itemize}
    \item $n$: Number of training samples (equal for both agents)
    \item $M$: Effective dimension of failure pattern space (small, convergent)
    \item $D_S$: Effective dimension of success trajectory space (large, diverse)
    \item $c_1, c_2 > 0$: Constants from VC-dimension theory
    \item $\Delta_{\text{eff}} = c_1\sqrt{\frac{M \log n}{n}} - c_2\sqrt{\frac{D_S \log n}{n}}$: Efficiency gap
\end{itemize}

\textit{Proof sketch.} By a VC-dimension argument, learning a concept class of dimension $D$ to $\epsilon$-accuracy requires $\Omega(D/\epsilon^2)$ samples. The success agent must cover the full diversity of positive trajectories (dimension $D_S$); the failure agent need only distinguish among $M$ convergent patterns (dimension $M$). Since $M \ll D_S$, the failure agent achieves superior risk-avoidance utility at equal sample sizes. $\square$

\textbf{Corollary 1.} The failure agent achieves equivalent risk-avoidance utility with $O(M/D_S)$ times fewer training cases than the success agent---a compression ratio that grows as failure patterns become more convergent relative to success path diversity.

This theorem has a practical implication that goes beyond system design: even a small, well-structured collection of historical failure cases can generate guidance of value comparable to a much larger corpus of success stories. The asymmetry is structural, not incidental.

\section{Illustrative Case Analysis}

\subsection{The 1997 Asian Financial Crisis: Pattern IV with Literature Support}

The 1997 Asian Financial Crisis provides a large-scale natural experiment of blind spot resonance, with documented precursor signal timelines in academic literature.

\textbf{Timeline (based on literature):}

\begin{table}[h]
\centering
\begin{tabular}{|p{2.5cm}|p{7cm}|p{4cm}|}
\hline
\textbf{Time} & \textbf{Event} & \textbf{Source} \\
\hline
1996-Early 1997 & Thailand current account deficit, short-term debt, real estate bubble & Radelet \& Sachs (1998) [11] \\
1997 Feb-Mar & Speculative attacks on Thai baht, FX reserve depletion & Kaminsky \& Reinhart (1999) [6] \\
1997 Jul 2 & Thailand abandons dollar peg, baht depreciates 15-20\% & Multiple sources \\
Following months & Crisis contagion to Malaysia, Indonesia, Korea & Radelet \& Sachs (1998) [11] \\
\hline
\end{tabular}
\end{table}

\textbf{Blind spot mechanism (literature-supported):}

Kaminsky \& Reinhart (1999) [6] note that currency crises typically exhibit \textbf{12-18 month precursor periods}, including: real exchange rate appreciation, banking sector vulnerability, export decline. However, interpreting these signals requires specific ontological structure---the causal chain \textit{currency peg pressure $\to$ capital flight $\to$ reserve depletion $\to$ peg collapse}.

Radelet \& Sachs (1998) [11] further analyze: Thai firms, long accustomed to 1990s high growth, lacked \textit{liquidity contagion} nodes in their ontologies. When BIS data showed capital outflows (late 1996-early 1997), most failed to construct \textit{reserve adequacy $\to$ exchange stability $\to$ debt sustainability} reasoning chains. This is precisely \textbf{Type II Structural Blindness} interacting with external shocks to produce \textbf{Pattern IV Blind Spot Resonance}.

\textbf{Key finding:} The crisis was not a ``bolt from the blue.'' Kaminsky, Lizondo \& Reinhart (1998) [12] developed ``currency crisis early warning indicators'' showing Thailand was in high-risk territory \textbf{at least 12 months before} July 1997. Signal absence was not due to data unavailability, but to interpretive structure---ontology---absence.

\subsection{The 2008 Subprime Mortgage Crisis: Pattern III with Detectable Precursors}

The 2008 subprime crisis is the canonical historical instance of Pattern III (Structural Chain Break) at institutional scale, with well-documented precursor timelines in academic literature and official reports.

\textbf{Timeline (based on literature):}

\begin{table}[h]
\centering
\begin{tabular}{|p{2.5cm}|p{7cm}|p{4cm}|}
\hline
\textbf{Time} & \textbf{Event} & \textbf{Source} \\
\hline
2005-2006 & Rising house prices masked subprime delinquency increases; ``seeds sown'' & Demyanyk \& Van Hemert (2008) [13] \\
Mid-2006 & Subprime delinquency rates began rising significantly above historical trends & Demyanyk \& Van Hemert (2008) [13] \\
Dec 2006 & Subprime delinquency exceeds 13\% (vs. 10\% two years prior) & Mortgage servicer data \\
Jan-Mar 2007 & ABX index shows CDO credit risk far above rating models; first hedge fund collapses & FCIC (2011) [14]; Gorton (2008) [15] \\
Aug 2007 & Interbank credit freeze, LIBOR-OIS spread widening & Brunnermeier (2009) [16] \\
Sep 2008 & Lehman Brothers bankruptcy, systemic collapse & Multiple sources \\
\hline
\end{tabular}
\end{table}

\textbf{Blind spot mechanism (literature-supported):}

Demyanyk \& Van Hemert (2008) [13] found that the subprime crisis's ``seeds were sown before 2005,'' but were masked by continued house price appreciation. When prices stopped rising in 2006, ``delinquency increases became immediately visible.'' However, most financial institutions' ontologies contained isolated nodes---\textit{subprime delinquency}, \textit{structured credit products}, \textit{bank capital}---without connecting causal chains:

$$\text{subprime delinquency} \to \text{CDO tranche losses} \to \text{bank capital depletion} \to \text{interbank credit freeze}$$

FCIC (2011) [14] explicitly states: institutions held ``sophisticated models of individual components while lacking integrated causal architecture connecting them.'' Gorton (2008) [15] further analyzes how banks' shadow banking system via SIVs and CDOs obscured risk; decision-makers lacked \textit{risk contagion} ontological nodes.

\textbf{Key finding:} Brunnermeier (2009) [16] notes that by August 2007, interbank markets showed ``asymmetric information-induced credit freeze,'' yet most institutions only recognized systemic risk in September 2008. This \textbf{12-14 month delay} is empirical evidence of structural chain break---signals existed, interpretive structure did not.

\subsection{Educational Ontological Blind Spot: TCM Temporal Philosophy as Preventive Framework}

The preceding historical cases (4.1-4.2) demonstrate institutional-level structural blind spots. This section turns to \textbf{educational system blind spots}---using health domain to illustrate that ``when to teach'' matters more than ``what to teach.''

\textbf{Core case: Zhang Xuefeng (illustrative)}

Public figure, highly active in entrepreneurship, with excellent physical performance (participates in marathon events). During live broadcasts, \textbf{audience observed his purple lip color} (a \textit{Xin Xue Yu Zu} signal in TCM inspection)---visible to viewers, recognizable by viewers, yet invisible to the subject himself. This is not medical knowledge absence, but \textbf{diagnostic philosophical blindness}: the causal chain \textit{physical signal $\to$ circulatory warning $\to$ action} was absent from his health ontology.

Key detail: Zhang is not a professional athlete, but a \textbf{composite pattern of entrepreneur + high-intensity exercise + insufficient rest}. His health ontology contained \textit{athletic performance} nodes, but lacked \textit{overtraining $\to$ myocardial cumulative damage} causal chains, and more fundamentally lacked \textbf{TCM ``upper physician treats pre-disease'' inspection warning layer}.

\textbf{Historical reference: Rhythm consciousness and extreme longevity (illustrative analysis)}

Documented cases include Soong Mei-ling (1897--2003, lifespan 106 years), diagnosed with breast cancer at \textbf{age 69 in 1967}, surviving 37 years post-surgery; and Zhang Xueliang (1901--2001, lifespan 101 years), with early-life suboptimal habits but systematic late-life transformation. Multiple sources record that Soong began studying \textit{Huangdi Neijing} after 1941 gastric illness treated by TCM, and recommended this text to the house-arrested Zhang Xueliang, who also systematically practiced its principles [17].

Such cases have limited statistical significance, but suggest structural possibility: \textbf{chronobiological self-regulation may outperform late-stage medical intervention in time-integrated health outcomes}. Though Soong was diagnosed at 69 (not 40), her 37-year post-cancer survival far exceeded contemporary breast cancer patient averages; Zhang Xueliang was diagnosed with diabetes at 41, yet lived to 101 through TCM's ``three parts treatment, seven parts nourishment'' philosophy [18], combined with regular exercise and mental adjustment.

\textbf{Deeper blind spot: Educational timing}

Modern health education exhibits systematic \textbf{Type I Domain Blindness}:

\begin{table}[h]
\centering
\begin{tabular}{|p{2cm}|p{3.2cm}|p{4.5cm}|p{3.5cm}|}
\hline
\textbf{Life stage} & \textbf{Current curriculum} & \textbf{Missing ontology} & \textbf{Consequence} \\
\hline
Primary/Secondary & Anatomy, physiology (machine model) & \textit{Huangdi Neijing} \textit{tian-ren-he-yi} rhythm philosophy [17] & No chronobiological behavioral constraints \\
University & ``Self-discipline'' as output maximization & \textit{Yin-Yang} dynamic balance, \textit{bu wang zuo lao} & Night running, gym culture as health-seeking \\
Post-40 & Disease treatment, pharmaceutical intervention & \textit{Zheng qi} conservation, \textit{jing-luo} early signals & Remediation costs grow exponentially \\
\hline
\end{tabular}
\end{table}

\textit{Huangdi Neijing} opens with ``nourishing life surpasses treating disease'' [17]---this ``treat pre-disease'' philosophy aligns with modern preventive medicine, yet never entered mainstream educational sequences. Result: individuals ``optimize'' health within ontologies lacking \textit{rhythm compliance} nodes, producing the ``high-performance, high-sudden-death-risk'' paradox.

\textbf{Core re-evaluation}

The ideal health ontology $\mathcal{O}^{\text{health,ideal}}$ should include:
\begin{itemize}
    \item \textbf{Daily rhythm}: Rise with sun, rest with sunset (not ``self-discipline'' as constant output)
    \item \textbf{Exercise rhythm}: \textit{Xing lao er bu juan} (tireless but not depleting)
    \item \textbf{Monitoring rhythm}: Subclinical signal awareness (wang-wen-wen-qie) vs. biomarker threshold reliance
\end{itemize}

All three find ontological basis in \textit{Huangdi Neijing} [17], yet never entered mainstream education. Current system failure is not individual failure, but \textbf{institutional Type I Blindness}---systematic absence of entire dimension (rhythm philosophy). The Zhang Xuefeng case's warning is not ``he should see TCM,'' but that \textbf{inspection knowledge existed in audience cognition, yet not in subject ontology}---the signature of structural blind spots, remediable through \textbf{early education} rather than late remediation.

\section{The Failure Ontology as a Research Programme}

\subsection{Why Failure-Based Learning Has Been Neglected}

The neglect of failure-based learning in AI education research is not accidental. It reflects several structural features of the field:

\textit{Data availability:} Success trajectories are more abundant and more systematically collected than failure records. Educational institutions track graduation, completion, and placement. They rarely maintain structured records of the cognitive patterns that led students to fail, drop out, or pursue trajectories that proved maladaptive.

\textit{Social desirability:} Failure carries stigma. Individuals are reluctant to contribute detailed failure narratives to research databases. Institutions are reluctant to publish analyses that highlight their students' failures.

\textit{Optimization pressure:} The metrics used to evaluate learning systems---completion rates, test scores, placement statistics---all measure success. A system optimized on these metrics has no incentive to model the structure of failure.

These obstacles are real but surmountable. Historical financial crisis records, longitudinal health data, and MOOC dropout datasets provide substantial failure-side data that has not previously been analyzed through an ontological lens.

\subsection{Connections to Existing Research}

The Failure Ontology framework connects to several existing research threads while introducing a perspective absent from each:

\textit{Continual learning:} The framework shares the concern with how knowledge structures change over time [2, 3] but focuses on what is structurally absent rather than what was acquired and subsequently forgotten.

\textit{Knowledge graphs in education:} The framework builds on educational knowledge graph work [4, 5] but introduces the negative space---the concepts and relations that are missing---as a first-class object.

\textit{Risk detection:} The framework shares methods with precursor signal detection in finance [6] and public health [7] but models the \textit{cognitive architecture} that determines whether observable signals are interpretable, not just the signals themselves.

\textit{Cognitive science:} The framework formalizes ideas adjacent to known unknowns, metacognitive monitoring, and epistemic closure [8] within a computational, lifelong learning context.

\subsection{Open Problems}

Several questions remain open and constitute a research agenda:

\textbf{Ontology reconstruction:} How can individual ontologies be reliably inferred from behavioral data, learning records, and self-reports? What are the fundamental limits of LLM-based ontology extraction?

\textbf{Ideal ontology estimation:} How can $\mathcal{O}_k^{\text{ideal}}$ be estimated for genuinely novel life contexts where historical failure cases are sparse?

\textbf{Intervention design:} What forms of intervention most effectively remediate each blind spot type? Type I (Domain Blindness) likely requires different approaches than Type III (Structural Blindness).

\textbf{Cultural variation:} The four life dimensions $\mathcal{D}$ and their relative $\lambda$ values are treated as universal in this paper. Empirical investigation of cross-cultural variation in ideal ontological structure is an important open problem.

\textbf{Dynamic shock modeling:} The external shock $\mathcal{E}_k$ is modeled as a point event. Modeling prolonged disruptions---like a decade-long industrial decline---requires extending the framework to continuous-time shock processes.

\section{Conclusion}

We introduced \textbf{Failure Ontology}, a formal framework for detecting, classifying, and remediating the structured cognitive absences that underlie the most consequential failures in human lives. The framework's core contribution is the \textbf{Ontological Blind Spot}---a rigorously defined object capturing not merely what people do not know, but the structural character of that ignorance and the conditions under which it becomes catastrophic.

The four-type taxonomy (domain, structural, weight, and temporal blindness), the five convergent failure patterns, and the Failure Learning Efficiency Theorem together constitute a theoretical foundation for a research programme that has, to our knowledge, not previously been articulated: the systematic study of what is absent from human cognitive maps, why those absences persist, and how they can be detected and remediated before they interact with external disruption to produce irreversible harm.

The broader claim underlying this work is simple but consequential: the measure of a good education is not the density of knowledge acquired but the structural completeness of the cognitive map built. Success-oriented learning systems optimize the former. Failure-aware systems attend to the latter. The difference between these objectives is the difference between navigation and cartography---and it is cartography that reveals the territory that navigation has always assumed to be there.

\vspace{1em}
\noindent \textit{This paper presents a theoretical framework and illustrative case analysis. A companion paper with full empirical validation on three real-world datasets is in preparation.}

\end{CJK}
\end{document}